\documentclass[runningheads]{llncs}
\usepackage{balance}
\usepackage{amsmath}
\usepackage[tight,footnotesize]{subfigure}
\usepackage{bm}
\usepackage{algorithm}
\usepackage[noend]{algorithmic}
\usepackage{multirow}
\usepackage{amsfonts}

\DeclareMathOperator*{\argmax}{arg\,max}

\begin{document}




\title{Partial-Monotone Adaptive Submodular Maximization}
%
%
\author{Shaojie Tang\inst{1}  \orcidID{0000-0001-9261-5210} \and Jing Yuan\inst{2} \orcidID{0000-0001-6407-834X}}
\authorrunning{S. Tang and J. Yuan}
%
\institute{Naveen Jindal School of Management, University of Texas at Dallas\\\email{shaojie.tang@utdallas.edu}\and Department of Computer Science and Engineering, University of North Texas \\\email{jing.yuan@unt.edu}}

\maketitle              
\begin{abstract}Many AI/Machine learning problems require adaptively selecting a sequence of items, each selected item might provide some feedback that is valuable for making better selections in the future, with the goal of maximizing an adaptive submodular function. Most of existing studies in this field focus on either monotone case or non-monotone case. Specifically, if the utility function is monotone and adaptive submodular, \cite{golovin2011adaptive} developed $(1-1/e)$ approximation solution subject to a cardinality constraint. For the cardinality-constrained non-monotone case, \cite{tang2021beyond} showed that a random greedy policy attains an approximation ratio of $1/e$. In this work, we generalize the above mentioned results by studying the partial-monotone adaptive submodular maximization problem. To this end, we introduce the notation of adaptive monotonicity ratio $m\in[0,1]$ to measure the degree of monotonicity of a function. Our main result is to show that for the case of cardinality constraints, if the utility function has an adaptive monotonicity ratio of $m$ and it is adaptive submodular, then a random greedy policy attains an approximation ratio of $m(1-1/e)+(1-m)(1/e)$. Notably this result recovers the aforementioned $(1-1/e)$ and $1/e$  approximation ratios when $m = 1$ and $m = 0$, respectively. We further extend our results to consider a knapsack constraint and develop a $(m+1)/10$ approximation solution for this general case. One important implication of our results is that even for a non-monotone utility function, we still can attain an approximation ratio close to  $(1-1/e)$ if this function is ``close'' to a monotone function. This leads to improved performance bounds for many machine learning applications whose utility functions are almost adaptive monotone.
\end{abstract}


\section{Introduction}
We consider an adaptive stochastic optimization problem whose input is a set of items, each of which has a random state. The realized state of an item can not be observed until it has been selected. We assume there is a utility function whose value is jointly decided by the set of selected items and the realization of all items' states.  Our goal is to design an adaptive policy that builds up a set adaptively to maximize the utility.   That is, after selecting an item, we observe its realized state, which can be leveraged to make better decisions in the future.   Consider experiment design as a running example, in this application, each item represents a test such as temperature, and the state of an item represents a possible outcome from a test. For example, possible states of a temperature test could be \emph{high}, \emph{low}, and \emph{normal}. Clearly, one is not able to observe the realized state of a test until we select that test. In this example, our objective is to come up with a sequence of tests to draw an accurate conclusion about a patient. Given that finding efficient policies for this problem is hard in general, we focus on a class of utility functions called adaptive submodular function \cite{golovin2011adaptive}. The notation of adaptive submodularity is an adaptive version of the classic notation of submodularity, and it can be found in many practical problems such as active learning and adaptive influence maximization.

We note that most of existing results on maximizing an adaptive submodular function  \cite{chen2013near,tang2020influence,tang2020price,yuan2017adaptive,fujii2019beyond,tang2021partial,fairness,doi:10.1287/ijoc.2022.1239,tang2021optimal,tang2021adaptive,tang2021non} often assume that this function is either non-monotone or monotone. For the monoton case, a simple greedy policy attains a tight $1-1/e$ approximation ratio subject to a cardinality constraint \cite{golovin2011adaptive}. For the non-monotone case, \cite{tang2021beyond} developed a $1/e$ approximation solution.
In general, monotone objective functions admit improvement performance bounds as compared with general non-monotone functions. Observing that in many machine learning applications, their objective functions are close to monotone functions, this raises the following question: Can we derive improved approximation ratios for those \emph{near} monotone functions? Before answering this question, we first introduce the notation of  \emph{adaptive monotonicity ratio} $m\in[0,1]$ to measure the degree of adaptive monotonicity of a function. 
Intuitively,  an adaptive monotonicity ratio of $m=1$ corresponds to an adaptive monotone function, and lower values of $m$ indicate some violation of adaptive monotonicity. Our main results are twofold:
\begin{enumerate}
\item For maximizing a cardinality-constrained $m$-adaptive monotone and adaptive submodular function, we show that a simple random greedy policy attains an approximation ratio of $m(1-1/e)+(1-m)(1/e)$ against the optimal adaptive policy. Note that if we set $m=1$ and $m=0$, then our results recover \cite{golovin2011adaptive}'s results and \cite{tang2021beyond}'s results respectively.
\item For the case of   general knapsack constraints, we develop a $ \frac{m+1}{10}$ approximation solution. This recovers \cite{tang2021beyond1}'s results for the non-monotone case if we set $m=0$.
\end{enumerate}

We give a summary of studies related to ours in Table \ref{rrr}.

\begin{table}[t]
\begin{center}
\begin{tabular}{ |c|c|c|c| }
\hline
Source & Approximation ratio & Constraint &  Adaptive monotonicity ratio  \\
\hline
\cite{golovin2011adaptive} & $1-\frac{1}{e}$ &cardinality  & $m=1$ (monotone)\\
\cite{tang2021beyond} & $\frac{1}{e}$ &cardinality  & $m=0$ (non-monotone) \\
this work & $m(1-\frac{1}{e})+(1-m)\frac{1}{e}$ &cardinality  & general $m$ \\
\hline
\cite{tang2021beyond1}& $\frac{1}{10}$ & knapsack  & $m=0$ (non-monotone) \\
this work & $\frac{m+1}{10}$ & knapsack  & general $m$ \\
\hline
\end{tabular}
\caption{Summary of related studies}
\label{rrr}
\end{center}
\end{table}

\emph{ Additional related work.} 
In the field of traditional submodular maximization, \cite{iyer2015submodular} relaxes the assumption of monotonicity and  introduced the concept of monotonicity ratio, a continuous version of monotonicity. \cite{mualem2022using} provides a systematical study about this property. We extend this notation from sets to policies to provide enhanced performance bounds of several existing polices if the utility function is nearly adaptive monotone.
\section{Preliminaries}

\subsection{Items, States and Policies} Assume we are given a set  $E$ of $n$ items and each item $e\in E$ has a random state $\Phi(e)\in O$ through a function $\Phi: E\rightarrow O$, where $O$ is the state space for a single item.  In the example of experiment design, $E$ represents all possible tests, e.g., $E=\{\verb"temperature test", \verb"blood pressure test"\}$ and $O=\{\verb"high", \verb"normal", \verb"low"\}$, and the state   $\Phi(e)$  of each test $e\in E$ represents the outcome of $e$, e.g., \[\Phi(\verb"temperature test")\in O.\] We use $\phi: E\rightarrow O$  to represent a \emph{realization} of $\Phi$. Hence, for each item $e\in E$, $\phi(e)\in O$ is the realization of $\Phi(e)$. The realization of $\Phi(e)$ is not known until $e$ has been selected. We assume that the distribution of $\Phi$ is known, that is, we know the probability $p(\phi)=\Pr[\Phi=\phi]$ of each realization  $\phi\in U$, where $U$ state space for all items. A \emph{partial realization} $\psi\subseteq E\times O$  represents  the realizations of any subset of items, and we call this subset the \emph{domain} of $\psi$ (denoted by $\mathrm{dom}(\psi)$). Hence, $\phi$ is a partial realization with $\mathrm{dom}(\phi)=E$. We call a realization $\phi$ consistent with a partial realization $\psi$ (denoted  by $\phi \sim \psi$), if $\phi(e)=\psi(e)$ for all $e\in \mathrm{dom}(\psi)$.

We define an adaptive policy as a function $\pi: 2^E\times O^E \rightarrow \mathcal{P}(E)$ from partial
realizations (e.g., the current observation) to a distribution of $E$ (e.g., the next item to select). For the example of experiment design, $\pi$ would specify which test to perform next, given the outcomes from past tests. 

\begin{definition}
Given two policies $\pi$ and $\pi'$,  we define $\pi @\pi'$ as a new policy that runs $\pi$ first, and then runs $\pi'$ from a fresh start.
\end{definition}

In addition, we assume there is a utility function   $f: 2^E\times O^E \rightarrow \mathbb{R}_{\geq0}$. For any subset of items $A$ and any realization $\phi$, the value of $f(A, \phi)$ measures the utility of selecting $A$ conditional on $\phi$. We  define the expected  utility of  a policy $\pi$ as follows
\begin{equation}
f_{avg}(\pi)=\mathbb{E}_{\Phi\sim p(\phi), \Pi}f(E(\pi, \Phi), \Phi),
\end{equation}
where $E(\pi, \phi)$ denotes the set of items selected by $\pi$ given $\phi$; $\Pi$ represents the internal randomness of a policy $\pi$.

\subsection{Problem Formulation}

Assume selecting an item $e\in E$ incurs a cost $c(e)$, our objective is to find a policy  $\pi^{opt}$ to maximize the expected utility subject to a budget constraint $k$:
\[\pi^{opt} \in \argmax_{\pi:\forall \phi\in U^+, \sum_{e\in E(\pi, \phi)}c(e)\leq k} f_{avg}(\pi),\]
where $U^+=\{\phi\mid p(\phi)>0\}$ represents the set of \emph{possible} realizations.

That is, we aim to build up a set $A$, whose total cost is at most $k$, sequentially and adaptively such that $f(A,\phi)$ is maximized where $\phi$ is unknown initially.

\subsection{Additional Notations}
\begin{definition}
\label{def:1}
The marginal utility  of adding an item $e$ to $\psi$ is
\[\Delta(e \mid \psi)=\mathbb{E}_{\Phi}[f(\mathrm{dom}(\psi)\cup \{e\}, \Phi)-f(\mathrm{dom}(\psi), \Phi)\mid \Phi \sim \psi].\]
Similarly, the marginal utility of adding a set of items $S\subseteq E$ to $\psi$ is
\[\Delta(S \mid \psi)=\mathbb{E}_{\Phi}[f(\mathrm{dom}(\psi)\cup S, \Phi)-f(\mathrm{dom}(\psi), \Phi)\mid \Phi \sim \psi].\]
\end{definition}

\begin{definition}
\label{def:12}
The marginal utility of running a policy $\pi$ on top of $\psi$ is
\[\Delta(\pi\mid \psi)=\mathbb{E}_{\Phi}[f(\mathrm{dom}(\psi) \cup E(\pi, \Phi), \Phi)-f(\mathrm{dom}(\psi), \Phi)\mid \Phi\sim \psi].\]
\end{definition}

With these notations, we are in position to introduce two important concepts from \cite{golovin2011adaptive}.
\begin{definition}
\label{def:11}
A function  $f$ is called \emph{adaptive submodular}, if
\[\Delta(e\mid \psi) \geq \Delta(e\mid \psi') \]  for all $\psi\subseteq \psi'$ and all $e\in E\setminus \mathrm{dom}(\psi')$.
\end{definition}

\begin{definition}
\label{def:2}
A function  $f$ is called \emph{adaptive monotone}, if
\[\Delta(e\mid \psi) \geq 0 \] for all  $\psi$ and all  $e\in E$.
\end{definition}

We next introduce the notation of \emph{adaptive monotonicity ratio} $m\in[0,1]$. Intuitively, this ratio, which can be viewed as an adaptive version of the (non-adaptive) monotonicity ratio introduced in \cite{iyer2015submodular,mualem2022using}, captures the degree of adaptive monotonicity of a function $f$.
\begin{definition}
The adaptive monotonicity ratio of a function  $f$  is
\begin{eqnarray}
m\overset{\Delta}{=} \min_{\pi, \pi'} \frac{f_{avg}(\pi@\pi')}{f_{avg}(\pi)}
\end{eqnarray}
where the ratio $\frac{f_{avg}(\pi@\pi')}{f_{avg}(\pi)}$ is assumed to be $1$ if $f_{avg}(\pi) = 0$.
\end{definition}

As proved in \cite{golovin2011adaptive}, if $f$ is adaptive monotone, then $f_{avg}(\pi@\pi')\geq f_{avg}(\pi)$ for any two policies $\pi$ and $\pi'$. Hence, $ \frac{f_{avg}(\pi@\pi')}{f_{avg}(\pi)}\geq 1$ where the equality occurs when $\pi'$ does not select any items, i.e., $ \frac{f_{avg}(\pi@\pi')}{f_{avg}(\pi)}= 1$ if $\forall \phi\in U^+$, $ E(\pi', \phi)=\emptyset$. This implies that if $f$ is adaptive monotone, then its adaptive monotonicity ratio is one, i.e.,  $\min_{\pi, \pi'} \frac{f_{avg}(\pi@\pi')}{f_{avg}(\pi)}=1$.

\section{Cardinality Constraint}
We first study a special case of our problem by assuming that each item has a unit cost, i.e., $\forall e\in E, c(e)=1$. Assuming $k$ is an integer, our problem is reduced to a cardinality-constrained optimization problem:
\[\max_{\pi:\forall \phi\in U^+, |E(\pi, \phi)| \leq k} f_{avg}(\pi).\]

We introduce a  random greedy policy, called \emph{Adaptive Random Greedy Policy} (labeled as $\pi^{arg}$) to solve this problem. This algorithm was originally proposed in \cite{gotovos2015non,tang2021beyond}. The detailed pseudocode of running $\pi^{arg}$ in listed in Algorithm \ref{alg:LPP1}. In the initialization stage, we expand the ground set $E$ by adding  a set $D$ of $2k-1$ dummy items, such that, $\forall \psi, \forall d\in D, \Delta(d \mid \psi) =0$. Let $E'=E\cup D$. The purpose of adding these dummy items is to make sure we always select items with non-negative marginal utility. Moreover, because the marginal utility of every dummy item is zero, removing any dummy items from the output does not affect its utility. Now we are ready to present $\pi^{arg}$. It starts with an empty set and an empty partial realization $\psi^0=\emptyset$. In each subsequent round $r\in[k]$ where $[k]$ represents $\{0, 1, 2, \cdots, k\}$, $\pi^{arg}$  randomly selects an
item $e_r$ from the set $M(\psi^{r-1})$, where
 $M(\psi^{r-1})$ contains the items with $k$ largest gains on top of  $\psi^{r-1}$,  i.e., $M(\psi^{r-1})\leftarrow \arg\max_{M\subseteq E'; |M|\leq k} \sum_{e\in E'}\Delta(e\mid \psi^{r-1})$, here $\psi^{r-1}$ is the partial realization observed before entering round $r$. After observing $\Phi(e_r)$,  $\pi^{arg}$  updates the partial realization as follows:  $\psi^r= \psi^{r-1}\cup\{(e_r, \Phi(e_r))\}$, and enters the next round. This procedure iterates until we select $k$ (possibly dummy) items. We obtain the final output by removing all dummy items from the solution.

\begin{algorithm}[hptb]
\caption{Adaptive Random Greedy Policy $\pi^{arg}$ \cite{gotovos2015non,tang2021beyond}}
\label{alg:LPP1}
\begin{algorithmic}[1]
\STATE $A=\emptyset; r=1; \psi^0=\emptyset$.
\WHILE {$r \leq k$}
\STATE compute $M(\psi^{r-1})\leftarrow \arg\max_{M\subseteq E'; |M|\leq k} \sum_{e\in E'}\Delta(e\mid \psi^{r-1})$;
\STATE sample $e_r$ uniformly at random from $M(\psi^{r-1})$ and observe its realization $\Phi(e_r)$;
\STATE update the partial realization: $\psi^r= \psi^{r-1}\cup\{(e_r, \Phi(e_r))\}$;
\STATE $A\leftarrow A\cup \{e_r\}$; $r\leftarrow r+1$;
\ENDWHILE
\RETURN $A$ (after removing all dummy items)
\end{algorithmic}
\end{algorithm}


 It has been shown that $\pi^{arg}$  achieves a tight approximation ratio of $1-1/e$ for the monotone case \cite{gotovos2015non}. For the non-monotone case, \cite{tang2021beyond} show that $\pi^{arg}$ attains a $1/e$ approximation ratio. In Theorem \ref{thm:1}, we generalize their results by showing that $\pi^{arg}$  achieves an approximation ratio of $m(1-1/e)+(1-m)(1/e)$ given that the adaptive monotonicity ratio of $f$ is $m$. Note that if we set $m=1$ and $m=0$, then our results recover \cite{gotovos2015non}'s results and \cite{tang2021beyond}'s results respectively.
\begin{theorem}
\label{thm:1}
If $f$ is adaptive submodular and $m$-adaptive monotone with respect to $p(\phi)$, then the Adaptive Random Greedy Policy $\pi^{arg}$ achieves an approximation ratio of $m(1-1/e)+(1-m)(1/e)$ in expectation  using $O(n k)$ value oracle queries.
\end{theorem}

\emph{Proof:} The running time of  $\pi^{arg}$ follows from the observation that $\pi^{arg}$ takes $k$ rounds and each round takes $O(n)$ value oracle queries. We next focus on proving the approximation ratio of $\pi^{arg}$. For every $r\in [k]$, define $\pi^{arg}_r$ as a policy that runs  $\pi^{arg}$ for $r$ rounds. We first provide a technical lemma to bound the distance between $f_{avg}(\pi^{opt}@\pi^{arg}_r) $ and $f_{avg}(\pi^{opt})$.

\begin{lemma}
\label{lem:1111}If $f$ is adaptive submodular and $m$-adaptive monotone with respect to $p(\phi)$, then for every $r\in [k]$,
\begin{equation*}
f_{avg}(\pi^{opt}@\pi^{arg}_r) \geq m  f_{avg}(\pi^{opt}) + (1-m)(1-\frac{1}{k})^{r}f_{avg}(\pi^{opt}).
\end{equation*}
\end{lemma}
\emph{Proof:} Let $(\psi^{opt}, \psi^{r-1})$ be a partial realization after running $\pi^{opt}@\pi^{arg}_{r-1}$, where $\psi^{opt}$ is the partial realization after running $\pi^{opt}$ and $\psi^{r-1}$ is the partial realization after running $\pi^{arg}_{r-1}$. Let $\psi^{o@arg_{(r-1)}}= \psi^{opt}\cup\psi^{r-1}$.  Consider a fixed $(\psi^{opt}, \psi^{r-1})$,
\begin{eqnarray}
&&\mathbb{E}_{e_r}[\Delta(e_r \mid\psi^{o@arg_{(r-1)}})]~\nonumber\\
&=& \frac{1}{k} \sum_{e\in M(\psi^{r-1})} \Delta(e \mid\psi^{o@arg_{(r-1)}})~\nonumber\\
&\geq& \frac{1}{k}\Delta(M(\psi^{r-1}) \mid \psi^{o@arg_{(r-1)}})]~\nonumber\\
&=& \frac{1}{k}\bigg(\mathbb{E}_{\Phi}[f_{avg}(\mathrm{dom}(\psi^{o@arg_{(r-1)}})\cup M(\psi^{r-1}), \Phi)\mid\Phi\sim \psi^{o@arg_{(r-1)}}] ~\nonumber\\ &&\quad\quad-\mathbb{E}_{\Phi}[f(\mathrm{dom}(\psi^{o@arg_{(r-1)}}), \Phi)\mid \Phi\sim \psi^{o@arg_{(r-1)}}]\bigg)\label{eq:ddd}
\end{eqnarray} where  the first equality is by the rule of selecting $e_r$ and the inequality is by the assumption that $f$ is adaptive submodular. Let $\Psi^{o@arg_{(r-1)}}= \Psi^{opt}\cup\Psi^{r-1}$ where $\Psi^{opt}$ is a random realization of $\psi^{opt}$ and $\Psi^{r-1}$ is a random realization of $\psi^{r-1}$. Let $\pi^{arg+}_{r-1}$ denote a policy that selects the first $r-1$ items using $\pi^{arg}_{r-1}$, then adds all items from $ M(\Psi^{r-1})$ to the solution. Hence,
\begin{eqnarray}
&&\mathbb{E}_{(\Psi^{opt}, \Psi^{r-1})}\bigg[\mathbb{E}_{\Phi}[f_{avg}(\mathrm{dom}(\Psi^{o@arg_{(r-1)}})\cup M(\Psi^{r-1}), \Phi)\mid\Phi\sim \Psi^{o@arg_{(r-1)}}]\bigg]~\nonumber\\
&&= f_{avg}(\pi^{opt}@\pi^{arg+}_{r-1}).\label{eq:sunday}
\end{eqnarray}

Unfixing  $(\psi^{opt}, \psi^{r-1})$, taking the expectation of $\mathbb{E}_{e_r}[\Delta(e_r \mid\Psi^{o@arg_{(r-1)}})]$ over $(\Psi^{opt}, \Psi^{r-1})$, we have
\begin{eqnarray}
&&\mathbb{E}_{(\Psi^{opt}, \Psi^{r-1})}\left[\mathbb{E}_{e_r}[\Delta(e_r \mid\Psi^{o@arg_{(r-1)}})]\right] \\
&=& f_{avg}(\pi^{opt}@\pi^{arg}_{r})-f_{avg}(\pi^{opt}@\pi^{arg}_{r-1}) \\
&\geq&   \mathbb{E}_{ (\Psi^{opt}, \Psi^{r-1})}\bigg[\frac{1}{k}(\mathbb{E}_{\Phi}[f_{avg}(\mathrm{dom}(\Psi^{o@arg_{(r-1)}})\cup M(\Psi^{r-1}), \Phi)\mid\Phi\sim \Psi^{o@arg_{(r-1)}}] ~\nonumber\\
&& -\mathbb{E}_{\Phi}[f(\mathrm{dom}(\Psi^{o@arg_{(r-1)}}), \Phi)\mid \Phi\sim \Psi^{o@arg_{(r-1)}}])\bigg]\\
&=& \frac{1}{k} (f_{avg}(\pi^{opt}@\pi^{arg+}_{r-1})-f_{avg}(\pi^{opt}@\pi^{arg}_{r-1}))\\
&\geq& \frac{1}{k} (m f_{avg}(\pi^{opt})-f_{avg}(\pi^{opt}@\pi^{arg}_{r-1})) \label{eq:9999}
\end{eqnarray}
where the first inequality is by (\ref{eq:ddd}), the second equality is by (\ref{eq:sunday}), and the second inequality is by the assumption that $f$ is  $m$-adaptive monotone.

Now we are in position to prove this lemma by induction on the number of rounds $r$. For the base case when $r=0$, this lemma is true because $f_{avg}(\pi^{opt}@\pi^{arg}_0) =  f_{avg}(\pi^{opt}) \geq m f_{avg}(\pi^{opt}) + (1-m)(1-\frac{1}{k})^{0}f_{avg}(\pi^{opt})$. Suppose this lemma is true for any $r'$ such that $r'<r$, we next prove it for $r>0$.
\begin{eqnarray*}
&&f_{avg}(\pi^{opt}@\pi^{arg}_r) = f_{avg}(\pi^{opt}@\pi^{arg}_{r-1})+ (f_{avg}(\pi^{opt}@\pi^{arg}_{r})-f_{avg}(\pi^{opt}@\pi^{arg}_{r-1}))\\
&\geq&  f_{avg}(\pi^{opt}@\pi^{arg}_{r-1})+ \frac{1}{k} (m f_{avg}(\pi^{opt})-f_{avg}(\pi^{opt}@\pi^{arg}_{r-1}))\\
&=& \frac{m}{k}  f_{avg}(\pi^{opt})+(1-\frac{1}{k})f_{avg}(\pi^{opt}@\pi^{arg}_{r-1})\\
&\geq & \frac{m}{k}  f_{avg}(\pi^{opt})+(1-\frac{1}{k})(m  f_{avg}(\pi^{opt}) + (1-m)(1-\frac{1}{k})^{r-1}f_{avg}(\pi^{opt}))\\
&=& m  f_{avg}(\pi^{opt}) + (1-m)(1-\frac{1}{k})^{r}f_{avg}(\pi^{opt}).
\end{eqnarray*} The first inequality is by (\ref{eq:9999}) and the second inequality is by the inductive assumption. $\Box$

We next focus on proving the theorem. Recall that  $\Psi^{r-1}$ is a random partial realization after running $\pi^{arg}_{r-1}$. The expectation $\mathbb{E}_{ \Psi^{r-1}}[\cdot]$ is taken over all such partial realizations $\Psi^{r-1}$. It follows that
\begin{eqnarray}
&&f_{avg}(\pi^{arg}_r) - f_{avg}(\pi^{arg}_{r-1}) ~\nonumber\\
&=& \mathbb{E}_{ \Psi^{r-1}}[\mathbb{E}_{e_r}[\Delta(e_r \mid\Psi^{r-1})]]\nonumber\\
&=& \frac{1}{k} \mathbb{E}_{\Psi^{r-1}} [\sum_{e\in M(\Psi^{r-1})} \Delta(e \mid\Psi^{r-1})]\nonumber\\
&\geq& \frac{1}{k} \mathbb{E}_{ \Psi^{r-1}} [ \Delta(\pi^{opt} \mid\Psi^{r-1})]\nonumber\\
&=& \frac{f_{avg}(\pi^{opt}@\pi^{arg}_{r-1})-  f_{avg}(\pi^{arg}_{r-1})}{k}\nonumber\\
&\geq& \frac{m f_{avg}(\pi^{opt}) + (1-m)(1-\frac{1}{k})^{r}f_{avg}(\pi^{opt})-  f_{avg}(\pi^{arg}_{r-1})}{k}.\label{eq:bbbb}
\end{eqnarray}
The second equality is by the design of $\pi^{arg}$, the first inequality is by the assumption that $f$ is adaptive submodular and Lemma 1 in \cite{gotovos2015non}, and the second inequality is by Lemma \ref{lem:1111}.

 We next prove
 \begin{equation}\label{eq:xxx}
 f_{avg}(\pi^{arg}_{r})\geq \bigg(m(1-(1-\frac{1}{k})^r)+ (1-m)\frac{r}{k} (1-\frac{1}{k})^{r-1}\bigg)f_{avg}(\pi^{opt})
 \end{equation} by induction on the number of rounds $r$. This follows the same proof of Theorem 4.2 by \cite{mualem2022using}. For the base case when $r=0$,  $f_{avg}(\pi^{arg}_{0})\geq 0 \geq \bigg(m(1-(1-\frac{1}{k})^0)+ (1-m)\frac{0}{k} (1-\frac{1}{k})^{0-1}\bigg)f_{avg}(\pi^{opt})$. Assume (\ref{eq:xxx}) is true for $r'<r$, we next
prove it for $r$.
\begin{eqnarray*}
 &&f_{avg}(\pi^{arg}_{r}) \\
 &&\geq  f_{avg}(\pi^{arg}_{r-1})+ \frac{m f_{avg}(\pi^{opt}) + (1-m)(1-\frac{1}{k})^{r}f_{avg}(\pi^{opt})-  f_{avg}(\pi^{arg}_{r-1})}{k}\\
 &&= (1-1/k)f_{avg}(\pi^{arg}_{r-1})+\frac{m f_{avg}(\pi^{opt}) + (1-m)(1-\frac{1}{k})^{r}f_{avg}(\pi^{opt})}{k}\\
  &&\geq (1-1/k)\cdot \big(m(1-(1-\frac{1}{k})^{r-1}) + (1-m)\frac{r-1}{k} (1-\frac{1}{k})^{r-2}\big)f_{avg}(\pi^{opt})\\
  &&\quad\quad+\frac{m f_{avg}(\pi^{opt}) + (1-m)(1-\frac{1}{k})^{r}f_{avg}(\pi^{opt})}{k}\\
  &&=  \bigg(m(1-(1-\frac{1}{k})^r)+ (1-m)\frac{r}{k} (1-\frac{1}{k})^{r-1}\bigg)f_{avg}(\pi^{opt}).
\end{eqnarray*}
The first inequality is by  (\ref{eq:bbbb}), the second inequality is by the inductive assumption.
When $r=k$, we have $ f_{avg}(\pi^{arg}) \geq \bigg(m(1-(1-\frac{1}{k})^k)+ (1-m)(1-\frac{1}{k})^{k-1}\bigg)f_{avg}(\pi^{opt})\geq \bigg(m(1-1/e)+(1-m)(1/e)\bigg)f_{avg}(\pi^{opt})$. $\Box$

\section{Knapsack Constraint}

We next study our problem subject to a general knapsack constraint $k$. We introduce a \emph{Sampling-based Adaptive Density-Greedy Policy} (labeled as $\pi^{sad}$). This policy was originally proposed in \cite{tang2021beyond1}, which itself is inspired by the sampling technique used in \cite{amanatidis2020fast}.  $\pi^{sad}$ is composed of two candidate policies, namely, $\pi^1$ and $\pi^2$.
\begin{itemize}
\item For simplicity, let $f(e)=\Delta(e\mid \emptyset)$.  $\pi^1$ always selects the best singleton $e^*$ that has  the largest expected utility, i.e., $e^*=\arg\max_{e\in E} f(e)$.
\item $\pi^2$  follows a density-greedy rule to select items. Specifically, it first samples a random set $S$ such that each item from $E$ is included in $S$ independently with probability $1/2$. Starting with round $r=1$, an initial partial realization $\psi_0=\emptyset$ and an initial budget $C \leftarrow k$, in each subsequent round $r$, $\pi^2$ selects an item $e_r$ with the largest ``benefit-to-cost'' ratio from a subset $F$ given the partial realization $\psi^{r-1}$, i.e.,
\[e_r\leftarrow \arg\max_{e \in F}\frac{\Delta(e\mid  \psi^{r-1})}{c(e)}\] where  $F=\{e\in S|  \Delta(e\mid  \psi^{r-1})>0 \mbox{ and } C\geq c(e) \}$. After observing the realization of  $\Phi(e_r)$, we update the observation using
$\psi^r \leftarrow \psi^{r-1}\cup\{(e_r, \Phi(e_r))\}$, and update $C$ as follows: $C\leftarrow C-c(e_r)$. This procedures iterates until $F$ becomes empty.
\end{itemize}

In our final policy $\pi^{sad}$, we randomly pick one policy from $\{\pi^1, \pi^2\}$ to run such that the probability of running $\pi^1$ is  $1/5$ and the probability of running $\pi^2$ is  $4/5$.  We list the detailed description of $\pi^{sad}$ in Algorithm \ref{alg:LPP2}.

\begin{algorithm}[t]
\caption{Sampling-based Adaptive Density-Greedy Policy $\pi^{sad}$ \cite{tang2021beyond1}}
\label{alg:LPP2}
\begin{algorithmic}[1]
\STATE $S=\emptyset$, $e^*=\arg\max_{e\in E} f(e)$, $r=1$, $\psi^0=\emptyset$, $C = k$.
\FOR {$e\in E$}
\STATE  let $r_e \sim \mathrm{Bernoulli}(1/2)$
\IF {$r_e=1$}
\STATE $S=S \cup\{e\}$
\ENDIF
\ENDFOR
\STATE Sample $r_0$ uniformly at random from $[0,1]$
\IF [Running  $\pi^1$]{$r_0\in[0,1/5)$}
\STATE pick $e^*$
\ELSE [Running $\pi^2$]

\STATE $F=\{e|e\in S, f(e)>0\}$
\WHILE {$F \neq \emptyset $}
\STATE select $e_r\leftarrow \arg\max_{e \in F}\frac{\Delta(e\mid  \psi^{r-1})}{c(e)}$ and observe  $\Phi(e_r)$
\STATE update the partial realization $\psi^r= \psi^{r-1}\cup\{(e_r, \Phi(e_r))\}$
\STATE $C=C-c(e_r)$
\STATE $S=S\setminus\{e_r\}$, $F=\{e\in S|  \Delta(e\mid  \psi^{r-1})>0 \mbox{ and } C\geq c(e) \}$, $r\leftarrow r+1$
\ENDWHILE
\ENDIF
\end{algorithmic}
\end{algorithm}

Before providing the main theorem, we prove three technical lemmas. The following corollary is adapted from \cite{tang2021beyond1} (Corollary 1 therein).
\begin{corollary}\cite{tang2021beyond1}\label{cor:1}
If $f$ is adaptive submodular with respect to $p(\phi)$, then \[4f_{avg}(\pi^2) + f(e^*) \geq  f_{avg}(\pi^{opt}@\pi^2).\]
\end{corollary}

Given a policy $\pi$, let $\mathrm{range}(\pi)=\{e | e\in \cup_{\phi\in U^+} E(\pi, \phi)\}$ denote the set of all items that has a positive probability  of being selected by $\pi$. The following is a key lemma for analyzing the performance bound of our solution.
\begin{lemma}
\label{lem:2}
If $f$ is adaptive submodular and $m$-adaptive monotone with respect to $p(\phi)$, then for any three policies $\pi^{a}$, $\pi^{b}$, and $\pi^{c}$ such that \[\mathrm{range}(\pi^{b})\cap \mathrm{range}(\pi^{c})=\emptyset,\] we have
\begin{eqnarray}
f_{avg}(\pi^a@ \pi^b) + f_{avg}(\pi^a @\pi^c)\geq  (1+m)f_{avg}(\pi^a).
\end{eqnarray}
\end{lemma}
\emph{Proof:}
We first present a useful inequality from \cite{tang2021beyond1} (Inequality (5) therein)  for any three policies $\pi^{a}$, $\pi^{b}$, and $\pi^{c}$ such that $\mathrm{range}(\pi^{b})\cap \mathrm{range}(\pi^{c})=\emptyset$. \begin{eqnarray}
&&f_{avg}(\pi^a@ \pi^b @\pi^c)~\nonumber\\
&&\leq f_{avg}(\pi^a)+(f_{avg}(\pi^a@ \pi^b)- f_{avg}(\pi^a)) + (f_{avg}(\pi^a@ \pi^c)- f_{avg}(\pi^a)).\label{eq:4}
\end{eqnarray}
Because $f$ is  $m$-adaptive monotone, we have $f_{avg}(\pi^a@ \pi^b @\pi^c)\geq m f_{avg}(\pi^a)$. This, together with (\ref{eq:4}), implies that
\begin{eqnarray}
&&f_{avg}(\pi^a)+(f_{avg}(\pi^a@ \pi^b)- f_{avg}(\pi^a)) + (f_{avg}(\pi^a@ \pi^c)- f_{avg}(\pi^a))~\nonumber\\
&&\geq m f_{avg}(\pi^a). \label{eq:5}
\end{eqnarray}
It follows that
 \begin{eqnarray*}
&&f_{avg}(\pi^a@ \pi^b) + f_{avg}(\pi^a @\pi^c)\\
&&=f_{avg}(\pi^a)+(f_{avg}(\pi^a@ \pi^b)- f_{avg}(\pi^a))\\
&&\quad\quad\quad+ f_{avg}(\pi^a) + (f_{avg}(\pi^a@ \pi^c)- f_{avg}(\pi^a))\\
&&\geq (1+m) f_{avg}(\pi^a).
\end{eqnarray*}
The inequality is due to (\ref{eq:5}). $\Box$

\begin{lemma}
\label{lem:11}If $f$ is adaptive submodular and $m$-adaptive monotone with respect to $p(\phi)$, then $f_{avg}(\pi^{opt}@\pi^2)\geq \frac{m+1}{2}f_{avg}(\pi^{opt})$.
\end{lemma}
\emph{Proof:} Recall that $\pi^2$ selects items from a random set $S$ in a density-greedy manner.  We next build a new policy $\pi^{2'}$  such that  $\pi^{2'}$ follows the same density-greedy rule to select items from $E\setminus S$. Hence, we can conclude that given a fixed partition $(S, E\setminus S)$, $\pi^2$ and $\pi^{2'}$ select items from two \emph{disjoint} subsets, that is, $\mathrm{range}(\pi^2)\cap \mathrm{range}(\pi^{2'})=\emptyset$ conditional on  any given $(S, E\setminus S)$. Letting $\mathbb{E}[f_{avg}(\pi^{opt}@\pi^2)+f_{avg}(\pi^{opt}@\pi^{2'})| (S, E\setminus S)]$ denote the conditional expected value of $f_{avg}(\pi^{opt}@\pi^2)+f_{avg}(\pi^{opt}@\pi^{2'})$ conditioned on a partition $(S, E\setminus S)$, Lemma \ref{lem:2} and the fact that $\mathrm{range}(\pi^2)\cap \mathrm{range}(\pi^{2'})=\emptyset$ imply that
\begin{eqnarray}
&&\mathbb{E}[f_{avg}(\pi^{opt}@\pi^2)+f_{avg}(\pi^{opt}@\pi^{2'})| (S, E\setminus S)]~\nonumber \\
&&\geq \mathbb{E}[(1+m)f_{avg}(\pi^{opt})| (S, E\setminus S)]\label{eq:89}
\end{eqnarray} for any  $S$.

Let $X_S$ denote a random variable of $S$. Now unfixing $(S, E\setminus S)$, taking the expectation of
\[\mathbb{E}[f_{avg}(\pi^{opt}@\pi^2)+f_{avg}(\pi^{opt}@\pi^{2'})| (X_S, E\setminus X_S)]\] over $(X_S, E\setminus X_S)$, we have
\begin{eqnarray}
&&\mathbb{E}_{(X_S, E\setminus X_S)}\bigg[\mathbb{E}[f_{avg}(\pi^{opt}@\pi^2)+f_{avg}(\pi^{opt}@\pi^{2'})| (X_S, E\setminus X_S)]\bigg]\\
&& \geq \mathbb{E}_{(X_S, E\setminus X_S)}\bigg[\mathbb{E}[(1+m)f_{avg}(\pi^{opt})| (X_S, E\setminus X_S)]\bigg]\\
&& =(1+m) f_{avg}(\pi^{opt}) \label{eq:6}
\end{eqnarray}
where the inequality is by (\ref{eq:89}). Note that \[\mathbb{E}_{(X_S, E\setminus X_S)}\bigg[\mathbb{E}[f_{avg}(\pi^{opt}@\pi^2)+f_{avg}(\pi^{opt}@\pi^{2'})| (X_S, E\setminus X_S)]\bigg]\]\[=f_{avg}(\pi^{opt}@\pi^2)+f_{avg}(\pi^{opt}@\pi^{2'}).\] This, together with (\ref{eq:6}), implies that
\begin{eqnarray}
f_{avg}(\pi^{opt}@\pi^2)+f_{avg}(\pi^{opt}@\pi^{2'})\geq(1+m) f_{avg}(\pi^{opt}). \label{eq:61}
 \end{eqnarray}

 Moreover, because $\pi^2$ and $\pi^{2'}$ are symmetric, we have $f_{avg}(\pi^{opt}@\pi^2)=f_{avg}(\pi^{opt}@\pi^{2'})$. This, together with (\ref{eq:61}), implies this lemma. $\Box$

We next  present the main theorem of this section.
\begin{theorem}If $f$ is adaptive submodular and $m$-adaptive monotone with respect to $p(\phi)$, then $f_{avg}(\pi^{sad})\geq  \frac{m+1}{10}f_{avg}(\pi^{opt})$.
\end{theorem}
\emph{Proof:}
Corollary \ref{cor:1} and Lemma \ref{lem:11} together imply that
\begin{eqnarray}
\label{eq:7}
4f_{avg}(\pi^1) + f(e^*) \geq \frac{m+1}{2}f_{avg}(\pi^{opt}).
\end{eqnarray}

Recall that  $\pi^{sad}$ picks a policy from $\{\pi^1, \pi^2\}$ to run such that the probability of selecting $\pi^1$ is  $1/5$ and the probability of selecting $\pi^2$ is  $4/5$. Hence, we can compute the expected utility of our final policy $\pi^{sad}$ as follows
\begin{eqnarray}
&&f_{avg}(\pi^{sad})= \frac{4}{5}\times f_{avg}(\pi^2) + \frac{1}{5}\times f(e^*)\\
&& =\frac{1}{5}\times (4f_{avg}(\pi^1) + f(e^*))\\
&& \geq \frac{m+1}{10}f_{avg}(\pi^{opt})
\end{eqnarray}
where the inequality is by (\ref{eq:7}).
$\Box$

\section{Conclusion}
In this paper, we study the partial monotone adaptive submodular maximization problem. We introduce the notation of adaptive monotonicity ratio to measure the degree of adaptive monotonicity of a function. We reanalyze the performance bound of several existing policies using this new notation. Our results show that a near monotone function enjoys improved performance bounds as compared with non-monotone functions.

\bibliographystyle{splncs04}
\bibliography{reference}

\begin{thebibliography}{10}
\providecommand{\url}[1]{\texttt{#1}}
\providecommand{\urlprefix}{URL }
\providecommand{\doi}[1]{https://doi.org/#1}

\bibitem{amanatidis2020fast}
Amanatidis, G., Fusco, F., Lazos, P., Leonardi, S., Reiffenh{\"a}user, R.: Fast
  adaptive non-monotone submodular maximization subject to a knapsack
  constraint. In: Advances in neural information processing systems (2020)

\bibitem{chen2013near}
Chen, Y., Krause, A.: Near-optimal batch mode active learning and adaptive
  submodular optimization. ICML (1)  \textbf{28}(160-168), ~8--1 (2013)

\bibitem{fujii2019beyond}
Fujii, K., Sakaue, S.: Beyond adaptive submodularity: Approximation guarantees
  of greedy policy with adaptive submodularity ratio. In: International
  Conference on Machine Learning. pp. 2042--2051 (2019)

\bibitem{golovin2011adaptive}
Golovin, D., Krause, A.: Adaptive submodularity: Theory and applications in
  active learning and stochastic optimization. Journal of Artificial
  Intelligence Research  \textbf{42},  427--486 (2011)

\bibitem{gotovos2015non}
Gotovos, A., Karbasi, A., Krause, A.: Non-monotone adaptive submodular
  maximization. In: Twenty-Fourth International Joint Conference on Artificial
  Intelligence (2015)

\bibitem{iyer2015submodular}
Iyer, R.K.: Submodular optimization and machine learning: Theoretical results,
  unifying and scalable algorithms, and applications. Ph.D. thesis (2015)

\bibitem{mualem2022using}
Mualem, L., Feldman, M.: Using partial monotonicity in submodular maximization.
  Advances in neural information processing systems  (2022)

\bibitem{tang2020price}
Tang, S.: Price of dependence: stochastic submodular maximization with
  dependent items. Journal of Combinatorial Optimization  \textbf{39}(2),
  305--314 (2020)

\bibitem{tang2021beyond}
Tang, S.: Beyond pointwise submodularity: Non-monotone adaptive submodular
  maximization in linear time. Theoretical Computer Science  \textbf{850},
  249--261 (2021)

\bibitem{tang2021beyond1}
Tang, S.: Beyond pointwise submodularity: Non-monotone adaptive submodular
  maximization subject to knapsack and k-system constraints. In: International
  Conference on Modelling, Computation and Optimization in Information Systems
  and Management Sciences. pp. 16--27. Springer (2021)

\bibitem{doi:10.1287/ijoc.2022.1239}
Tang, S.: Robust adaptive submodular maximization. INFORMS Journal on Computing
   (2022)

\bibitem{tang2020influence}
Tang, S., Yuan, J.: Influence maximization with partial feedback. Operations
  Research Letters  \textbf{48}(1),  24--28 (2020)

\bibitem{tang2021adaptive}
Tang, S., Yuan, J.: Adaptive regularized submodular maximization. In: 32nd
  International Symposium on Algorithms and Computation (ISAAC 2021). Schloss
  Dagstuhl-Leibniz-Zentrum f{\"u}r Informatik (2021)

\bibitem{tang2021non}
Tang, S., Yuan, J.: Non-monotone adaptive submodular meta-learning. In: SIAM
  Conference on Applied and Computational Discrete Algorithms (ACDA21). pp.
  57--65. SIAM (2021)

\bibitem{tang2021partial}
Tang, S., Yuan, J.: Partial-adaptive submodular maximization. arXiv preprint
  arXiv:2111.00986  (2021)

\bibitem{fairness}
Tang, S., Yuan, J.: Group equality in adaptive submodular maximization (2022).
  \doi{10.48550/ARXIV.2207.03364}, \url{https://arxiv.org/abs/2207.03364}

\bibitem{tang2021optimal}
Tang, S., Yuan, J.: Optimal sampling gaps for adaptive submodular maximization.
  In: AAAI (2022)

\bibitem{yuan2017adaptive}
Yuan, J., Tang, S.J.: Adaptive discount allocation in social networks. In:
  Proceedings of the 18th ACM International Symposium on Mobile Ad Hoc
  Networking and Computing. pp. 1--10 (2017)

\end{thebibliography}




\end{document}